%% file: main.tex
\documentclass[3p]{elsarticle}
\usepackage{lineno,hyperref}
\usepackage{amsmath}
\usepackage{amsfonts}
\usepackage[T1]{fontenc}

\usepackage{amsthm,array}
\usepackage{algorithm,algcompatible}
\usepackage{float}
\usepackage{subfigure}
\usepackage{algpseudocode}
\usepackage{algorithmicx}
\usepackage{changes} 
\usepackage{tikz} 
\usepackage{xcolor}
\usepackage{dsfont}
\usepackage{verbatim}
\usepackage{color}
\usepackage{url}
\usepackage{xurl}
\usepackage{soul}
\usepackage{booktabs}
\usepackage[title]{appendix}
\theoremstyle{plain}

\usepackage{graphicx}
\usepackage{caption}
\theoremstyle{definition}

\usepackage{arydshln} 
\usepackage{amssymb} 
\usepackage{multirow}
\usepackage{bm}
\theoremstyle{remark}

\modulolinenumbers[1]
\usepackage{setspace}
\setcitestyle{authoryear,round}
\usepackage[figuresright]{rotating}
\let\today\relax
\makeatletter
\def\ps@pprintTitle{%
    \let\@oddhead\@empty
    \let\@evenhead\@empty
    \def\@oddfoot{\footnotesize\itshape
         {Preprint} \hfill\today}%
    \let\@evenfoot\@oddfoot
    }
\makeatother

\journal{X}
\begin{document}
\begin{doublespace}
\begin{frontmatter}

\title{Data Mining in Transportation Networks with Graph Neural Networks:\\ A Review and Outlook}
\author[1]{Jiawei Xue}
\author[1]{Ruichen Tan}
\author[2]{Jianzhu Ma}
\author[1,3]{Satish V. Ukkusuri}

\address[1]{Lyles School of Civil and Construction Engineering, Purdue University, West Lafayette, IN, USA.}
\address[2]{Institute for AI Industry Research, Tsinghua University, Beijing, China.}
\address[3]{Corresponding Author: {\tt\small sukkusur@purdue.edu}}

\begin{abstract} 

Data mining in transportation networks (DMTNs) refers to using diverse types of spatio-temporal data for various transportation tasks, including pattern analysis, traffic prediction, and traffic controls. Graph neural networks (GNNs) are essential in many DMTN problems due to their capability to represent spatial correlations between entities. Between 2016 and 2024, the notable applications of GNNs in DMTNs have extended to multiple fields such as traffic prediction and operation. However, existing reviews have primarily focused on traffic prediction tasks. To fill this gap, this study provides a timely and insightful summary of GNNs in DMTNs, highlighting new progress in prediction and operation from academic and industry perspectives since 2023. First, we present and analyze various DMTN problems, followed by classical and recent GNN models. Second, we delve into key works in three areas: (1) traffic prediction, (2) traffic operation, and (3) industry involvement, such as Google Maps, Amap, and Baidu Maps. Along these directions, we discuss new research opportunities based on the significance of transportation problems and data availability. Finally, we compile resources such as data, code, and other learning materials to foster interdisciplinary communication. This review, driven by recent trends in GNNs in DMTN studies since 2023, could democratize abundant datasets and efficient GNN methods for various transportation problems including prediction and operation.

\end{abstract}
\end{frontmatter}
\input{introduction}

\input{problem}

\input{gnn}
\input{application_gnn}

\input{opportunities_gnn}

\input{data_code}

\input{conclusion}

\input{ack}

\bibliographystyle{plainnat}
\bibliography{ref}

\end{doublespace}
\end{document}

%% file: introduction.tex
\section{Introduction}
\label{intro}



In 2021, Google developed and globally deployed a graph neural network (GNN) model to predict estimated time of arrival (ETA) in transportation networks via Google Maps\footnote{https://deepmind.google/discover/blog/traffic-prediction-with-advanced-graph-neural-networks/}. Here, the ETA prediction denotes predicting ongoing duration of a trip along a specified route based on current road traffic conditions. This model demonstrated superior prediction performance compared to baseline models, with case studies in Los Angeles, New York, Singapore, and Tokyo. Furthermore, it achieved a relative reduction of negative ETA predictions by 16\% to 51\% across 19 cities in America, Europe, and Asia~\citep{derrow2021eta}. This project overcame research and production challenges, benefiting worldwide users in trip and route planning. As one of the most influential intelligent transportation applications in recent years, this project prompts three key questions that this review seeks to answer: 
\begin{itemize}
    \item \textbf{RQ1: What are GNNs and what advantages do they offer in modeling graph-structured data?} 
    \item \textbf{RQ2: What practical applications could GNNs provide in transportation networks?} 
    \item \textbf{RQ3: What are the promising research directions for GNNs in transportation networks?}
\end{itemize}


The ETA prediction is an instance of data mining in transportation networks (DMTNs), which involves extracting and harnessing valuable information from massive amounts of data within transportation networks. DMTNs contain the collection, processing, fusion, prediction, and operation using various types of traffic data. Specific examples include congestion propagation characterization~\citep{luan2022traffic}, standard traffic prediction tasks like traffic congestion prediction~\citep{rahman2023deep,feng2023macro,bogaerts2020graph,cui2020learning}, prediction of ride-hailing~\citep{ke2021joint} and E-scooters demand~\citep{song2023sparse}. Besides, DMTNs encompass traffic data imputation~\citep{chen2020nonconvex,nie2024imputeformer,nie2025collaborative}, crash risk analysis~\citep{zhao2024exploring}, system resilience evaluation~\citep{wang2020evaluation}, and vehicle route optimization~\citep{liu2022personalized}, etc. The DMTN tasks are essential for integrating the sensing and operation of transportation systems to enhance their utilities. For instance, through applications of traffic imputation and prediction techniques, online map navigation platforms can construct temporal traffic profiles, facilitating efficient traffic guidance for both private vehicles and public transit users. Despite its significance, many DMTN tasks can be difficult due to three primary factors: (1) the complexity of spatial traffic state relationships across distinct locations~\citep{wu2020learning,wu2021inductive,dstagnn}, (2) the dependency of transportation networks on human activities such as sport events~\citep{yao2021twitter}, and (3) the large number of nodes and edges in transportation networks in metropolitan areas~\citep{boeing2020multi}.

To address the above barriers, existing studies have utilized GNNs in various DMTN problems. Here, GNNs are advanced machine learning methods specifically designed for graph-structured data~\citep{manessi2020dynamic,velivckovic2023everything,corso2024graph}. These models integrate graph convolution operations with neural architectures, capturing internodal relationships along graph edges~\citep{scarselli2008graph,kipf2016semi,velivckovic2017graph,abu2019mixhop}. This property aligns seamlessly with the need to describe numerous entity-entity relationships, such as user-item interactions in recommendation systems~\citep{ying2018graph,chen2024macro}, protein-protein interactions in drug discovery~\citep{jimenez2020drug}, and atom-atom proximity in material exploration~\citep{merchant2023scaling}. In transportation networks, GNNs have driven innovations in modeling complicated interconnections between various types of spatial entities in DMTN problems~\citep{rahmani2023graph}. These include vehicles for intelligent drivings~\citep{chen2021graph}, sensors for traffic speed prediction~\citep{feng2023macro}, users for mobility action prediction~\citep{xue2024predicting}, road segments for travel time estimation~\citep{fang2020constgat}, origin-destination pairs for ride-sourcing services~\citep{ke2021predicting}, and airspace sites for air traffic density prediction~\citep{xu2023air}. The proliferation of seminal work on GNNs' applications in DMTNs calls for a systematic review and outlook in this domain.

\linespread{1.6}
\begin{table*}[ht]
\caption{\label{existing_review} Existing GNN reviews on general applications and specific areas.}
\centering
\resizebox{0.66\textwidth}{19.0mm}{
\begin{tabular}{@{}ll@{}}
\toprule[1.2pt]
\addlinespace[5pt]
Review & Scope \\ \addlinespace[4pt]
\hline
\addlinespace[5pt]
\cite{zhang2020deep}, \cite{zhou2020graph}, \cite{wu2020comprehensive}  &  \multirow{2}{*}{General}  \\
 \cite{keramatfar2022graph}, \cite{corso2024graph}   &   \\
\addlinespace[3pt]
 \cite{wu2022graph}, \cite{gao2023survey}  &  Recommendation systems\\
 \addlinespace[5pt]
 \cite{jin2023survey}  &  Time series mining \\
  \addlinespace[5pt]
 \cite{zhang2024expressive}  & Expressive power \\
   \addlinespace[5pt]
 \cite{lu2024survey} & Biology, finance, etc. \\
 \addlinespace[3pt]
 \bottomrule[1.2pt]
\end{tabular}}
\end{table*}

Several reviews have summarized fundamental GNN architectures and their variations across various fields (Table~\ref{existing_review}). For example, \cite{zhou2020graph} examined general GNN components, including graph convolutional networks~\citep{kipf2016semi} and gated GNNs~\citep{li2015gated}, as well as their variants. They also enumerated applications of these models in natural science, computer vision, and natural language processing up to 2020. Subsequent reviews on GNNs focused on specific domains, such as recommendation systems~\citep{wu2022graph} and time series analysis~\citep{jin2023survey}. Additionally, \cite{zhang2024expressive} examined the expressive power of GNNs, focusing on the influence of node indices on GNN outcomes. Focusing on industrial applications, \cite{lu2024survey} concluded the utilization of GNNs across diverse industrial domains, including biology and finance. Note that these reviews did not specifically focus on transportation networks.


\linespread{1.6}
\begin{table*}[h]
\caption{\label{table_compare} Comparison between this review and existing reviews about GNNs in transportation networks.}
\centering
\resizebox{0.85\textwidth}{24.0mm}{
\begin{tabular}{@{}llllll@{}}
\toprule[1.2pt]
\addlinespace[4pt]
\multirow{2}{*}{Review} & Review of & Review of & Details in & Outlook  & Data and\\
[0cm]
& traffic prediction & traffic operation & industry practices & after 2023 & code \\ 
\addlinespace[2pt] \hline
\addlinespace[4pt]
\cite{ye2020build} & $\surd$ &    &    &   &   \\
\addlinespace[4pt]
\cite{jiang2022graph} & $\surd$ &    &    &   &  $\surd$   \\
\addlinespace[4pt]
\cite{shaygan2022traffic} & $\surd$ &    &    &   &  $\surd$    \\
\addlinespace[4pt]
\cite{jin2023spatio}  & $\surd$ &    &   &  &  $\surd$  \\ 
\addlinespace[4pt]
\cite{rahmani2023graph} & $\surd$ & $\surd$  &  &   &  $\surd$    \\
\addlinespace[4pt]
\cite{wei2023guest} & $\surd$ & $\surd$   &   &  &    \\ 
\addlinespace[4pt]
\hline 
\addlinespace[4pt]
This review & $\surd$ & $\surd$  & $\surd$  &  $\surd$  & $\surd$ \\ 
\addlinespace[2pt]
\bottomrule[1.2pt]
\end{tabular}}
\end{table*}

Regarding GNNs in transportation networks, we listed existing literature reviews in Table~\ref{table_compare}. Specifically, \cite{shaygan2022traffic} discussed GNN methods for predicting traffic speed~\citep{li2017diffusion,guo2021hierarchical}, traffic flow~\citep{song2020spatial}, and both~\citep{zheng2020gman}. \cite{jiang2022graph} enumerated studies on predicting traffic flow and demand for various transportation modes such as railway, taxi, and bicycle. Nevertheless, the two reviews have not included studies on traffic operations like traffic signal control~\citep{devailly2021ig}. Subsequent reviews filled this gap by expanding GNN applications to traffic operations in intelligent transportation tasks~\citep{rahmani2023graph,wei2023guest}. However, they did not incorporate large-scale industry deployments of GNNs in transportation networks by digital services such as Google Maps~\citep{derrow2021eta}, Amap~\citep{dai2020hybrid}, and Baidu Maps~\citep{fang2020constgat}.

\begin{figure}[H]
    \centering
    \includegraphics[scale=0.70]{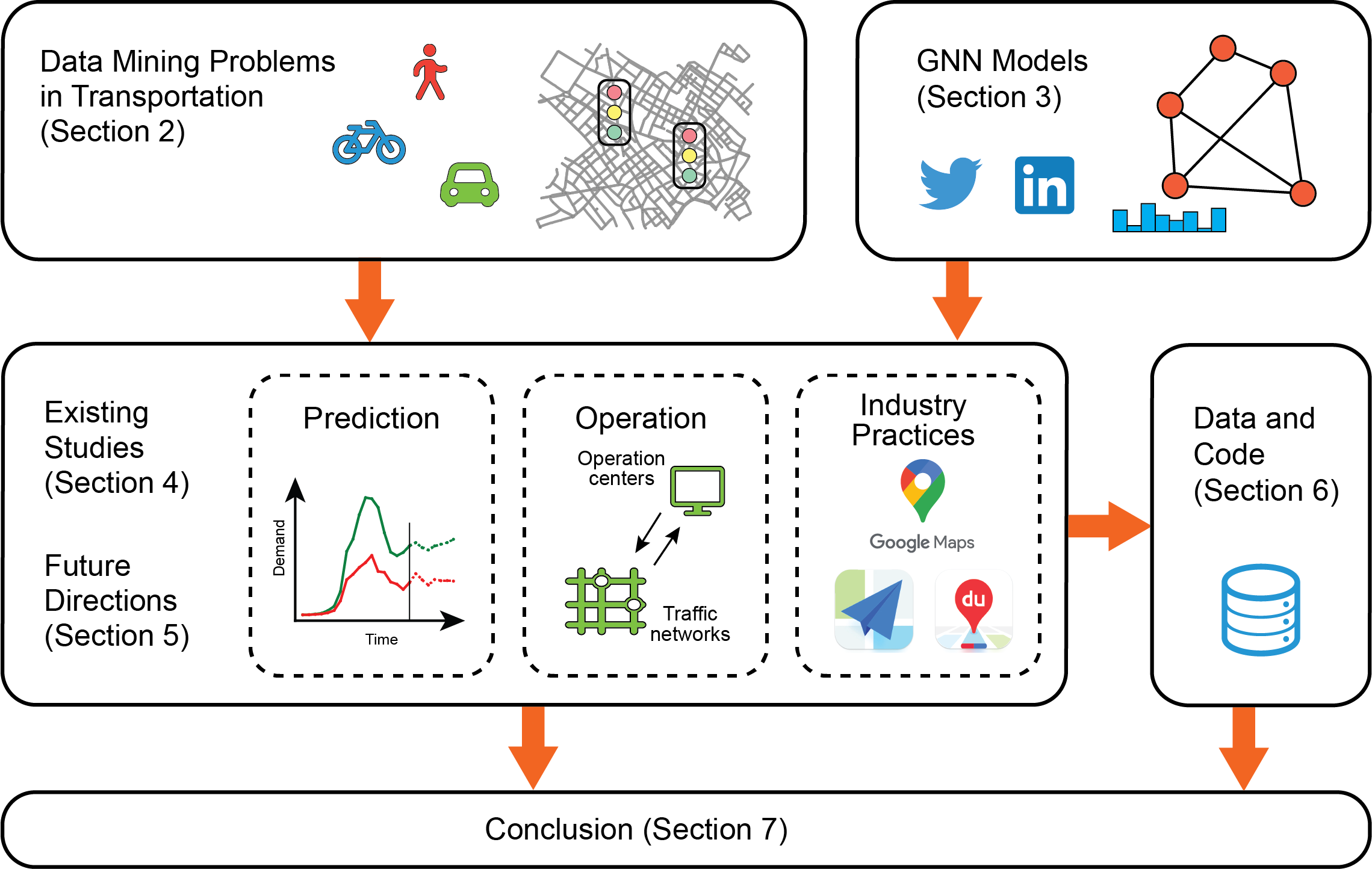}
    \caption{The framework of this survey. The survey begins by presenting data mining problems in transportation and various GNN models (Sections 2 and 3). Subsequently, we delve into current and prospective research on traffic prediction, operations, and industry-driven applications (Sections 4 and 5). Finally, we discuss data and code collections, followed by a conclusion (Sections 6 and 7).}
    \label{fig1}
\end{figure}

A thorough review addressing these drawbacks enables us to identify new research directions in mining and managing transportation systems. This review presents a comprehensive, up-to-date summary of GNN approaches used in DMTN problems, relevant to both academia and industry (Fig.~\ref{fig1}). First, we summarize critical DMTN problems including traffic prediction and operation. Second, we outline vanilla GNN models and their variants over time~\citep{you2020design}. Next, we analyze current applications and future opportunities of GNNs in DMTN problems. Finally, we highlight related online datasets, codes, and learning materials to support future endeavors for academic and industry readers. Together, our contributions are as follows.
\begin{itemize}
    \item We comprehensively analyze existing GNN studies in a wide spectrum of directions of traffic prediction and traffic operation across various transportation network components, including vehicles, sensor locations, road segments, and airspaces, from academic perspectives. Furthermore, we provide a detailed review of research progress and industry deployments by transportation services like Google Maps, highlighting their proprietary nature compared to academic approaches in problem definition and methodology development (Section~\ref{applications}).
    \item We discuss future GNN research directions, such as interval prediction, model simplifications, combinatorial problems, and traffic safety management, considering data availability and the suitability of methods (Section~\ref{opportunies}). 
    \item We categorize resources such as open datasets, codes, and tutorials for GNN methods and their applications in transportation networks, with an emphasis on the years 2023 and 2024 (Section~\ref{data}).
\end{itemize}

The following sections are scheduled as follows (Fig.~\ref{fig1}). Section~\ref{problem} outlines various DMTN problems, including traffic prediction and operation. Next, Section~\ref{gnn_model} discusses fundamental GNN models and their evolutionary variants. Section~\ref{applications} reviews current academic and industrial advances of GNNs in transportation networks. Section~\ref{opportunies} outlines future directions on new applications. Section~\ref{data} summarizes data, codes, and alternative resources to facilitate future study for readers. Finally, we conclude the review in Section~\ref{conclusion}.






%% file: problem.tex
\section{Prediction and Operation Problems in Transportation Networks}
\label{problem}

This section provides an overview of key DMTN problems amenable to GNNs. In particular, we review general formulations of traffic prediction, traffic operation in academic research, as well as travel time estimation in industrial applications.

\subsection{Traffic prediction}
\label{trafficprediction}
Traffic prediction refers to forecasting traffic variable values within transportation networks~\citep{wei2023guest,fafoutellis2023unlocking,yan2024imputation}. These networks can be modeled as graphs $G=(V, E)$, where $V$ and $E$ denote the sets of nodes and edges, respectively. In various transportation contexts, nodes and edges can represent different elements. In roadway networks with detected sensors, nodes represent sensor locations, while edges denote their proximities~\citep{mallick2020graph,ke2021joint,chen2024semantic}. In taxi and aircraft traffic networks, nodes typically represent regions and edges indicate traffic flows between these regions~\citep{yao2018deep,xu2023air}. In bike-sharing networks, nodes represent the locations of bike-sharing stations~\citep{cho2021enhancing}. Together, graphs provide a simple data structure for spatially representing the states and interactions of transportation entities, thereby bringing a wide range of applications in transportation.

The general traffic prediction problem can be defined as follows: given a transportation system with multiple entities (e.g., traffic stations, sensor locations, urban regions) and their traffic states by time $t$, predict these states after $t$. This formulation covers a wide variety of traffic prediction studies~\citep{li2017diffusion,yao2018deep,choi2022graph,rahman2023deep}. This task is crucial for providing future traffic conditions and demand estimations, which inform various traffic management policies. The key to an accurate traffic prediction model is to capture both the spatial and temporal dependencies inherent in traffic systems. Here, the temporal relationships can be appropriately modeled by time series methods, such as Gated recurrent units (GRUs)~\citep{chung2014empirical} and Transformers~\citep{wen2022transformers}. The spatial connections between traffic states at distinct locations can be effectively represented using GNNs. This is because GNNs can adaptively propagate information between nodes, which enables them to capture the complicated spatial traffic dependencies arising from traffic dynamics.

\subsection{Traffic operation}
Traffic operation includes a diverse range of management policies within transportation networks, like signal timing, vehicle operations, and transit management. In particular, we focus on two tasks where GNNs have been applied: vehicle routing~\citep{zhang2023route} and vehicle relocation~\citep{chang2022cooperative,lei2020efficient}. Vehicle routing is defined as follows: given origins, destinations, and operational constraints, determine optimal routes with specific goals such as cost minimization. Analogously, vehicle relocation involves the strategic redistribution of vehicles within urban road networks by selecting suitable destinations. Both vehicle routing and relocation are fundamental problems faced by individual drivers, transportation network companies like Uber~\citep{bertsimas2019online}, and logistics companies like FedEx~\citep{kitjacharoenchai2019vehicle}. 

\subsection{Industry practice}
A notable industrial application of GNNs is travel time estimation~\citep{derrow2021eta,dai2020hybrid,fang2020constgat}. The goal is to predict the future vehicle travel time along a route between a specified origin and destination. Practically, the total travel time for a route can be calculated by aggregating the travel times of road segments that constitute the route. Hence, the estimation of route travel time can be converted to the calculation of travel times for individual road segments. Travel time estimation holds immense value for individual travelers in their daily commutes and public agencies in their operational management. Individual travelers can decide the trip departure time based on expected arrival time and estimated travel time outcomes. Concurrently, road segment travel time serves as crucial input for public agencies to provide advance notifications of vehicle arrival time to passengers. This can significantly reduce passenger waiting time and enhance the user experience for public transit~\citep{li2024offline}.

%% file: gnn.tex
\section{Graph Neural Networks}
\label{gnn_model}

\subsection{Basic ideas}
\begin{figure}[H]
    \centering
    \includegraphics[scale=0.72]{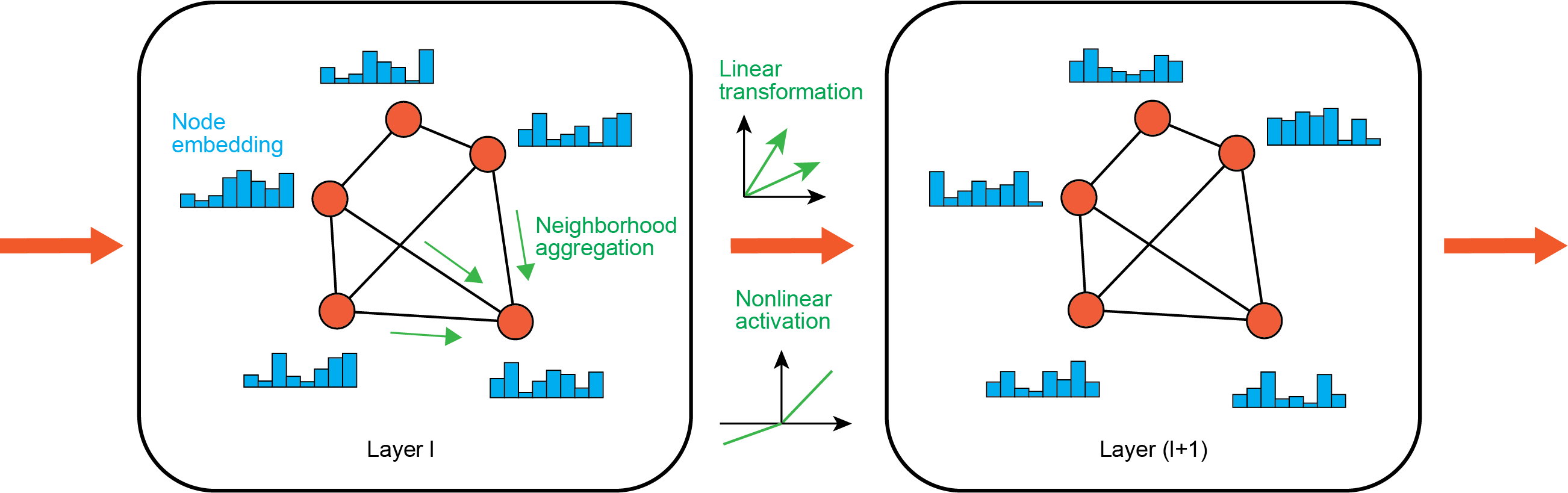}
    \caption{Three mechanisms in GNN models: neighborhood aggregation, linear transformation, and nonlinear activation. These mechanisms collectively transform node embeddings, which are numerical vectors associated with the nodes, from layer $l$ to layer $(l+1)$. Such information passing allows node embeddings to capture the graph's topology.}
    \label{gnn_mechanism}
\end{figure}

Graph-structured data, comprising entities and their interconnections, are prevalent across various disciplines~\citep{newman2018networks,hu2020open}. Mining patterns in such data can be challenging due to their complexity and the permutation-invariant property, meaning that the graph does not change after node reordering~\citep{keriven2019universal}. GNNs are specialized neural networks (NNs) designed to perform data mining by leveraging graph structure information~\citep{kipf2016semi,you2020design}. While classical NNs are composed of linear transformations and nonlinear activations (e.g., the sigmoid function, ReLU), GNNs employ a neighborhood aggregation mechanism to update node embeddings based on their neighborhoods (Fig.~\ref{gnn_mechanism}). This capability enables multiple tasks such as node classification, link prediction, and graph summarization~\citep{zhou2020graph}. 

Consider a graph with the node set $V$ ($|V|=n$) and the adjacency matrix $A$. Each node is represented by a $d$-dimensional embedding, forming the embedding matrix $H \in \mathbb{R}^{n\times d}$. A $k$-layer GNN can be expressed as a sequence of transformations between node embeddings:
\begin{equation}
    H^{l+1} = f(H^{l}, A, W), 0 \leq l \leq k-1.
    \label{eq1}
\end{equation}
Here, $H^{l}$ and $H^{l+1}$ denote node embeddings at layers $l$ and $(l+1)$. $W$ is the learnable weight matrix, and $f(\cdot)$ represents the mapping function. Specifically, $W$, $f(\cdot)$, and $A$ correspond to the linear transformation, nonlinear activation, and neighborhood aggregation modules, respectively. Note that Eq.~\ref{eq1} allows the final node embedding $H^{k}$ to capture the graph structure through neighborhood-based information fusion over $k$ iterations. 

\subsection{Generic GNN models}
As a specialized approach for processing graph-structured data, GNNs have undergone significant evolution, marked by numerous innovative architectures. This subsection provides a concise summary of the development of GNNs, highlighting key landmarks. The concept of GNNs was initially proposed by \cite{scarselli2008graph}. This groundbreaking work enabled the application of NNs to graph-structured data. The model achieved this by iteratively propagating information across nodes to reach a stable state. 

\cite{bruna2014spectral} proposed the spectral approach, utilizing the spectral decomposition of the graph Laplacian to learn about graph data. This work represented a giant advance in the field of GNNs. Following this, \cite{kipf2016semi} introduced graph convolutional networks (GCNs), which further streamlined the implementation of spectral graph convolution. This improvement enhanced the applicability of GNNs to large-scale graph data, particularly in node and graph classification tasks.

While GCNs require complete graph information simultaneously, graph attention networks (GATs) and GraphSAGE can aggregate neighborhood information locally. Here, GATs build attention mechanisms to dynamically determine the importance of information transfer between nodes, which is particularly effective in handling heterogeneous graph data~\citep{velivckovic2017graph}. GraphSAGE first fuses neighborhood representations and then concatenates them with the ego node's representation~\citep{hamilton2017inductive}. In 2018, graph isomorphism networks (GINs) were proposed in the paper \textit{How Powerful are GNNs} by \cite{xu2018powerful}. This work theoretically demonstrated the strong representation power of GNNs from the Weisfeiler-Lehman isomorphism test~\citep{huang2021short}, showing their ability to identify diverse graph structures.

Additionally, relational-GCN (R-GCN) accounts for the heterogeneity of edge types, supporting the modeling of graphs with diverse types of interactions~\citep{schlichtkrull2018modeling}. Furthermore, DiffPool aggregates node representations for each node cluster in a hierarchical manner, which contributes to the learning of complicated hierarchical graph structures~\citep{ying2018hierarchical}. More recently, \cite{kreuzer2021rethinking} introduced graph constraints into Transformer models~\citep{vaswani2017attention}, making another stride in the evolution of GNNs. This research employed learned positional encodings to capture node positional information and fed them into the Transformer structure.  

\subsection{GNN variants}
\label{gnn_variant}
\begin{figure}[h]
    \centering
    \includegraphics[scale=0.55]{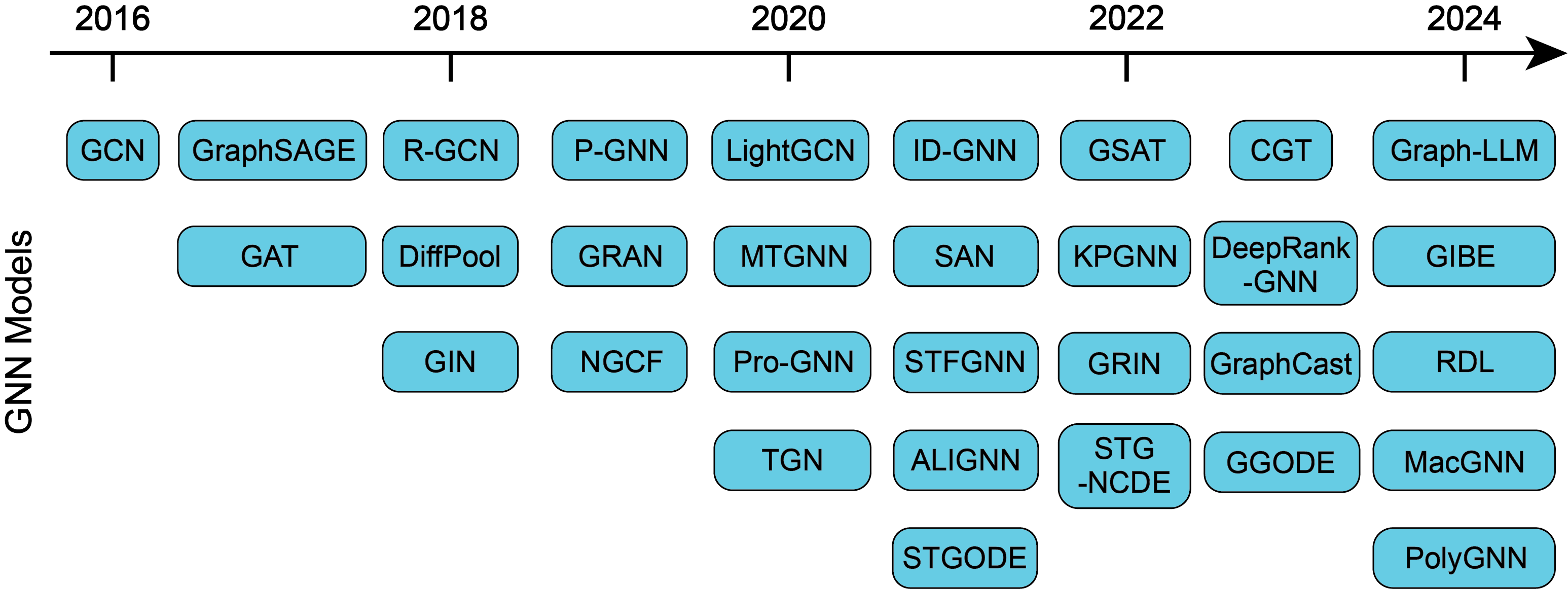}
    \caption{Prominent GNN models from 2016 to 2024. For models in the same year, the vertical positionings do not adhere to explicit criteria. Readers can navigate research articles for each model by referring to their model abbreviations.}
    \label{fig2}
\end{figure}

Following generic GNN models, the development of GNNs was driven by the universal properties of graph learning and a broad spectrum of applications (Fig.~\ref{fig2}). To overcome the universal limitations of generic GNNs, researchers have developed the following models. 
\begin{itemize}
    \item \textbf{Node positions}. Position-aware GNN (P-GNN) captures relative positions between an ego node and its neighborhood, thereby providing more rich node interconnection information~\citep{you2019position}. 
    \item \textbf{Large graphs}. To mitigate the computational complexity in large-scale graphs (e.g., those containing over one million nodes), researchers created either macro nodes by grouping original nodes~\citep{chen2024macro} or a series of multi-mesh nodes by summarizing multiple original nodes in local regions~\citep{lam2023learning}. In graph generation tasks, computation graph transformer (CGT) can capture the distribution of large-scale graphs while protecting user privacy~\citep{yoon2022graph}. 
    \item \textbf{Robust learning}. Given that GNNs are threatened by adversarial attacks like node perturbations, Property GNN was proposed to learn clear graph representations from noisy observations~\citep{jin2020graph}, by exploiting the low rank and sparsity characteristics. 
    \item \textbf{Out-of-distribution issues}. Out-of-distribution issues arise when a model deals with unobserved data that differs significantly from training data. In graph classification tasks, graph information bottleneck with explainability (GIBE) was developed to study the effect of regularization and its relationship with the out-of-distribution issues in GNNs~\citep{fang2024regularization}.
\end{itemize}

Furthermore, the widespread deployments of GNNs led to the development of domain-specific models, facilitating both the prediction and generation capabilities across various domains.  
\begin{itemize}
    \item \textbf{Recommendation systems}. Recommendation systems provide suggestions of products, locations, and services to users based on behavioral data~\citep{kang2018self,yan2024improving}. In user-item bipartite graphs, neural graph collaborative filtering (NGCF) propagates information between users and items along user-item edges~\citep{wang2019neural}. Later, researchers demonstrated the redundancy of linear transformations and nonlinear activation operations in NGCF~\citep{he2020lightgcn}. They introduced LightGCN, a streamlined yet effective GNN model, to predict user-item interactions in user check-in and review data.
    \item \textbf{Social networks}. Researchers from Twitter developed temporal graph networks (TGNs), which learn node representations from dynamic networks, to perform link prediction tasks in social networks from the Twitter platform~\citep{rossi2020temporal}. Another impactful deployment of GNNs in social networks was made by LinkedIn researchers~\citep{borisyuk2024lignn}. They integrated multiple entities (e.g., users, companies) and relations (e.g., posts, notifications) into a LinkedIn Graph with a hundred billion nodes. GNNs on the graph resulted in a 2$\%$ online improvement in advertisement clicks and a 1$\%$ increase in job applications.
    \item \textbf{Protein-protein interactions}. DeepRank-GNN can effectively capture protein-protein interactions to learn task-specific protein structure patterns relevant to drug design~\citep{reau2023deeprank}. Here, GNNs are invariant to node ordering, making them well-suited for modeling diverse interactions between proteins.
    \item \textbf{Integrated circuits}. In integrated circuit designs, a tree-based GNN model called SyncTREE facilitates timing prediction to support efficient timing updates throughout circuit design processes~\citep{hu2024synctree}.    
\end{itemize}

Readers can refer to alternative GNN models in reviews on recommendation systems~\citep{wu2022graph}, natural language processing~\citep{wu2023graph}, electrical engineering~\citep{chien2024opportunities}, and materials science~\citep{reiser2022graph}.

\subsection{Integration with large language models}

\begin{figure}[ht]
    \centering
    \includegraphics[scale=0.85]{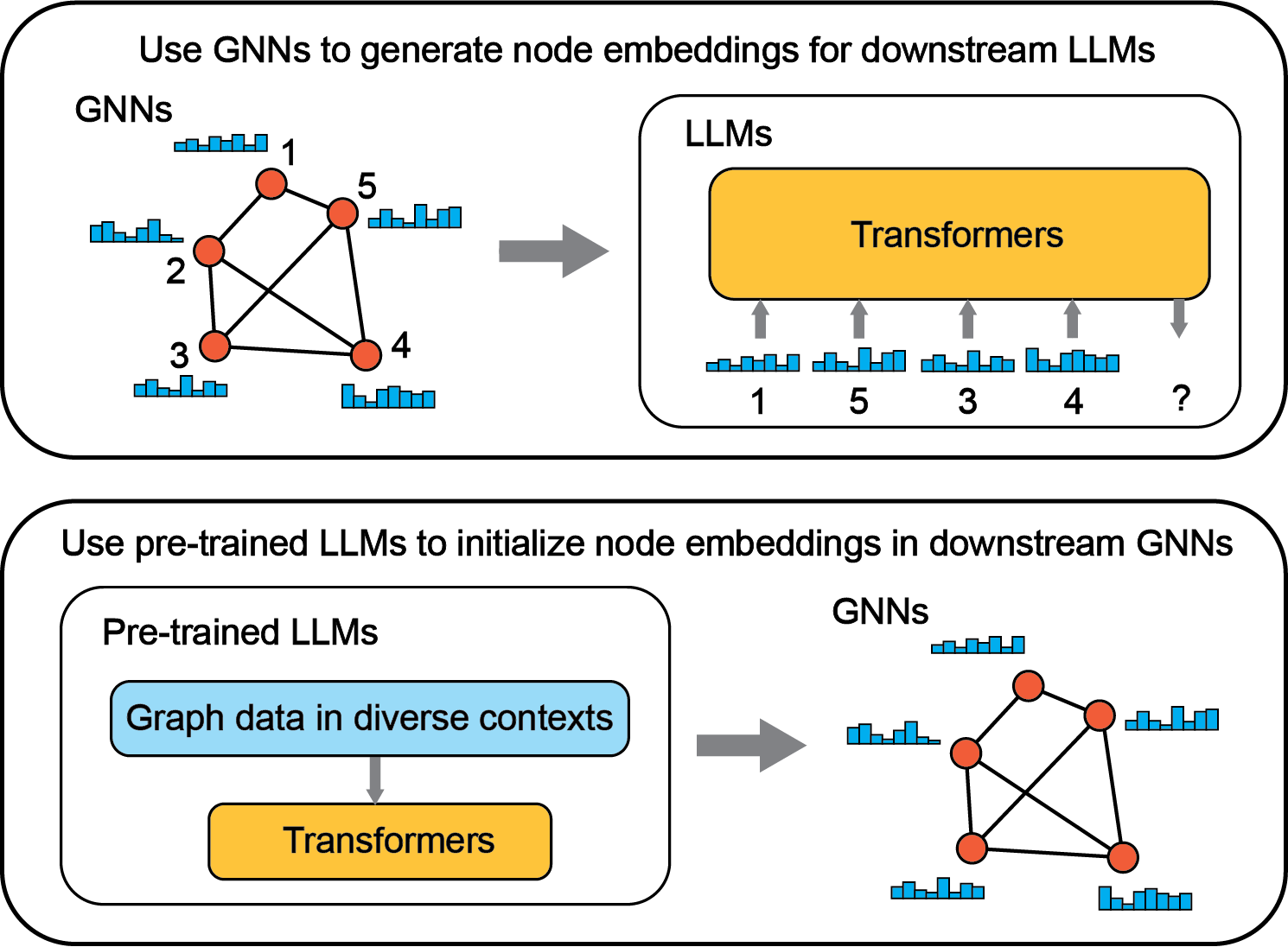}
    \caption{Two approaches to integrate GNNs and LLMs. The first approach employs GNNs to produce network topology-aware node embeddings and feeds them into afterward LLMs~\citep{perozzi2024let}. In contrast, the second approach leverages knowledge from pre-trained LLMs to enhance downstream GNNs~\citep{wei2024llmrec}, mitigating information deficiency during GNN training.}
    \label{fig_gnn_llm}
\end{figure}

GNNs are powerful for modeling graph-structured data but are vulnerable when node and edge information is incomplete. The recent proliferation of large language models (LLMs), including LLama2~\citep{touvron2023llama}, LLama3~\citep{dubey2024llama}, and GPT-4~\citep{achiam2023gpt}, offers immense potential for addressing this issue. In 2024, researchers have devised two primary approaches for integrating GNNs and LLMs (Fig.~\ref{fig_gnn_llm}). The first involves using GNNs to generate adjacency-aware node representations and tokens as inputs for downstream LLMs~\citep{ren2024survey,tsitsulin2024graph,perozzi2024let}. This structure leverages GNNs' strength in encoding non-Euclidean graph data, as well as LLMs' rich semantic knowledge. The second approach utilizes LLMs to generate node representations in graphs with incomplete information, providing more comprehensive features for both nodes and edges to support aftermath graph learning process~\citep{chen2024exploring}. One typical application involves enhancing sparse user-item graphs in recommendation systems with LLMs~\citep{wei2024llmrec}. These combinations of GNNs can simultaneously preserve the advantages of GNNs in representing relational structures within data and LLMs in extracting semantic information from sequential data.

\subsection{Interpretability of GNNs}
\label{general}
The interpretability of GNNs focuses on understanding the reasoning behind prediction outcomes, which is essential for real-world decision-making processes~\citep{ranu2024road,chen2024explainable}. Three key questions exist: given a trained GNN model, what features, subgraphs, and training samples significantly influence the prediction outcomes? These are categorized as feature-level, graph-level, and training data-level interpretation for GNNs, respectively.

First, multiple feature-level interpretation approaches have been introduced. For general NNs, SHapley Additive exPlanations (SHAP) assign each feature an importance value that satisfies desirable criteria~\citep{lundberg2017unified}. For GNNs, GNNExplainer achieves feature-level interpretability by identifying a subset of node features that yield prediction results comparable to those obtained using all node features~\citep{ying2019gnnexplainer}. To address the subset identification problem (equivalently, an integer program), the authors perform the relaxation using masked matrices with continuous values and optimize matrix values through backpropagation. Second, for graph-level interpretability, GNNExplainer identifies subgraphs that bring similar predictions for a given node to those of the complete graph. This is formulated as the maximization of mutual information which measures the predictive contribution of subgraphs. This technique has been employed to detect canner gene modules within biology research~\citep{li2024cgmega}. Another graph-level interpretation method XGNN focuses on generating subgraphs via reinforcement learning~\citep{yuan2020xgnn}.  Third, there is limited research focused on training data-level interpretation~\citep{yuan2022explainability}, primarily due to the extensive search space of training data subsets with diverse graph structures. Note that training data-level interpretability is critical as it enables navigating crucial training samples and reducing overall training costs~\citep{xie2023data}.


%% file: application_gnn.tex
\section{Current Applications of GNNs in Transportation Problems}
\label{applications}
\subsection{Traffic prediction}

As mentioned in Section~\ref{trafficprediction}, traffic prediction refers to forecasting traffic conditions (e.g., flow rate, traffic density, and average speed) within transportation networks. This process utilizes past traffic data and external factors, including weather~\citep{shaygan2022traffic} and special events~\citep{yao2021twitter}, as inputs and generates future traffic states as outputs. Initially, researchers applied standard feedforward NNs to traffic prediction tasks~\citep{ma2015long}. Subsequently, seminal GNN models, like GCNs, GraphSAGE, and GATs (discussed in Section~\ref{gnn_model}) catalyzed the development of various graph-based learning methods in traffic prediction, thanks to their advantages in capturing node relationships in networks~\citep{chen2023road,liang2023deep}. Notable models include DCRNN~\citep{li2017diffusion}, TGC-LSTM~\citep{cui2019traffic}, and AGC-Seq2Seq~\citep{zhang2019multistep}. The model proliferation led to comprehensive reviews by \cite{jiang2022graph} and \cite{rahmani2023graph}. While these studies and reviews predominantly focused on vehicular traffic speed and flow rate, recent research has expanded to cover other transportation modes, including bicycles~\citep{liang2023cross} and E-scooters~\citep{song2023sparse}. The latest methodologies also advance to novel model architectures, such as the macro-micro module, which unravels traffic patterns across multiple resolutions~\citep {feng2023macro}. This trend has been particularly apparent since 2023. This subsection reviews novel studies to elucidate these emerging directions. 

\subsubsection{Prediction targets}
As listed in Table~\ref{traffic_prediction}, GNNs have been continuously applied to predict traffic speed~\citep{ouyang2024tpgraph,feng2023macro} and flow~\citep{zou2024real,lv2023ts}.  Recent studies have also employed GNNs for aviation traffic prediction. For instance, \cite{xu2024unified} modeled specific airspace sectors as nodes, used the numbers of planes flying between two nodes to define adjacency matrices, and finally utilized GNNs to predict future aircraft density near airports. Similarly, \cite{li2024mast} predicted airspace complexity, categorizing it into three levels (low, normal, and high) within different sector regions, to aid air traffic management. Here, airspace complexity was determined by factors such as the sector volume, ground speed, and the number of planes. The applications of GNNs have expanded to forecasting traffic demand, including taxi traffic~\citep{tygesen2023unboxing}, metro passenger flow~\citep{li2023ig}, and bike sharing demand~\citep{liang2023cross}. These studies primarily focus on forecasting demand for existing infrastructure during its operation phase. In contrast, Metro-MGAT addresses the station ridership prediction problem for new metro stations during the planning phase~\citep{ding2024graph}.

Besides, \cite{cai2023fine} constructed causal graphs for pavement systems, and applied GNNs to these causal graphs to forecast pavement performance across the city. Together, the spatial modeling capability of GNNs facilitates the understanding of relationships between entities within networks, enabling a wide variety of traffic prediction tasks.

\begin{sidewaystable}[htp]
\caption{\label{traffic_prediction} Applications of GNNs in traffic prediction in 2023 and 2024. MPNN: Message Passing Neural Network. NYC: New York City. EV: Electric vehicle. For studies conducted before 2023, readers can consult existing reviews, including \cite{shaygan2022traffic} and \cite{rahmani2023graph}.}
\resizebox{1.00\textwidth}{65.5mm}{
\begin{tabular}{@{}llllllllr@{}}
 \toprule[1.5pt]
 \addlinespace[5pt]
Target  & Study & Model & Base module & Other modules  & Data  & Area  & Contribution   & Code \\
 \addlinespace[4pt]
 \toprule[1.5pt]
  \addlinespace[5pt]
Traffic speed, travel time & \cite{ouyang2024tpgraph} & TPGraph  & GCN  & Transformer   & Highway speed, travel time & California, Guizhou & Spatio-temporal & $\checkmark$ \\
 \addlinespace[3pt]
 Traffic speed  & \cite{jiang2023spatio} & MegaCRN & GCRN & Meta-graph learner  & Highway speed & California, Tokyo  & Spatio-temporal hetergeneity  & $\checkmark$\\
 \addlinespace[3pt]
Traffic speed  & \cite{feng2023macro} & MMSTNet & GCN & TCN, attention  & Highway speed & California  & Macro-micro  & $\checkmark$\\
 \addlinespace[3pt]
Traffic speed  & \cite{wang2023traffic}  & GSTAE & GCN  & GRU  & Highway speed  & California   & Handling missing values   & \\
 \addlinespace[3pt]
Traffic speed  & \cite{zhang2024adaptive}  & AIMST & GCN  & TCN  & Traffic speed  & Xi'an, Jinan   & Spatio-temporal, clustering  & \\
 \addlinespace[3pt]
 Traffic speed  & \cite{zhang2024context}  & CKG-GNN 
 & GNN  & Knowledge graphs  & Traffic speed  & Singapore   & Contextual information &  $\checkmark$\\
 \addlinespace[3pt]
 Traffic speed and flow  & \cite{ju2024cool} & COOL & MPNN & Attention & Highway speed and flow   & California  & High-order relationships &  \\
 \addlinespace[3pt]
Traffic speed and flow  & \cite{rahman2023deep} & DGCN-LSTM & GCN & LSTM & Highway speed and flow    & Florida  & Traffic during hurricane &  \\
 \addlinespace[3pt]
Traffic speed and flow  & \cite{ouyang2023domain} & DAGN & GCN & GRU & Traffic speed and flow  & California, Shenzhen  & Cross-city &  \\
 \addlinespace[3pt]
Transit flow  & \cite{zou2024real} & OD-PF & GCN  & Attention & Subway flow & Beijing   & OD during incidents & \\
 \addlinespace[3pt]
Traffic flow  & \cite{kong2024spatio} & STPGNN & Pivotal GCN   & Node identification & Taxi GPS, highway flow & California, Bejing, England & Pivotal nodes & \\
 \addlinespace[3pt]
Traffic flow  & \cite{chen2024traffic} & TFM-GCAM & GCN  & Attention & Highway flow & California   & Traffic flow matrix, Transformer & \\
 \addlinespace[3pt]
Traffic flow  & \cite{li2023spatial}  & STTGCN & GCN & Tensor, dilated conv & Highway flow  & California  & Binary adjacency matrix  & \\
 \addlinespace[3pt]
Traffic flow  & \cite{lv2023ts}  & TS-STNN  & Tree GCN & GRU & Highway flow & California  & Hierarchical and directional & \\ \addlinespace[3pt]
Congestion level  & \cite{feng2023urban}  & F-GCN    & GCN   & Attention, LSTM  & Taxi GPS & Beijing  & Congestion between segments  & \\ \addlinespace[3pt]
 \hline 
 \addlinespace[3pt]
Multi-traffic mode   & \cite{yang2024network} & M2-former  & GCN  & Attention& Subway, taxi, bus       & Beijing  & Multi-traffic  & \\  \addlinespace[3pt]
Metro demand  & \cite{ding2024graph} & Metro-MGAT & GAT  & Age-weighted loss & Metro ridership  & Shanghai  & Expanded demand  & \\
\addlinespace[3pt]
Metro demand  & \cite{li2023dynamic} & IG-Net & ChebNet  & Multi-task learning & Metro ridership  & Suzhou  & Three interactions  & \\
\addlinespace[3pt]
Traffic speed, taxi demand & \cite{tygesen2023unboxing} & NRI-X & Graph Network & VAE  & Highway speed, yellow taxi & NYC, California  & Optimal graphs  & $\checkmark$  \\  \addlinespace[3pt]
Charging demand & \cite{wang2023predicting}  & H-STGCN  & GCN  & GRU, clustering & EV trajectories  & Beijing   & Heterogeneous region scales  & \\  \addlinespace[3pt]
Bike sharing demand  & \cite{liang2023cross} & DA-MRGNN   & GGCN  & TCN     & Bike, subway, ride-hailing    & NYC   & Interactions between modes  & \\  \addlinespace[3pt]
E-scooter demand  & \cite{song2023sparse} & SpDCGRU  & Diffusion Conv  & GRU & E-scooter OD trips, etc.   & Louisville   & Spatio-temporal & \\  \addlinespace[3pt]
Taxi and bike demand   & \cite{chen2024semantic}  & SFMGTL   & ST-GNN & Clustering, etc.  & Taxi, bike & NYC, Chicago, Washington & Knowledge transfer & $\checkmark$ \\  \addlinespace[3pt]
Train delay   & \cite{huang2024explainable}  & GAT   & GAT & Attention  & Train, weather data & The Netherlands & Interpretability and accuracy &  \\  \addlinespace[3pt]
Ship trajectory  & \cite{zhang2023trajectory}  & G-STGAN   & GCN & Transformer, etc.  & Ship & Hong Kong, Weihai, Zhoushan & Integrates GCN and Transformer &  \\
\addlinespace[3pt]
 \hline 
 \addlinespace[3pt]
Air traffic    & \cite{xu2024unified}      & BEGAN    & GAT    & LSTM, Bayesian  & Flights, etc.    & Georgia, Flordia  & Knowledge-based    &                   \\
 \addlinespace[3pt]
Airspace complexity    & \cite{li2024mast}    & MAST-GNN  & GCN   & TCN-Att    & Air traffic & China   &  Spatio-temporal & $\checkmark$   \\
 \addlinespace[3pt]
 \hline 
 \addlinespace[3pt]
 Accident risk   & \cite{trirat2023mg}  & MG-TAR   & GCN & Correlations  & Accident risks, etc. & South Korea & Spatio-temporal &  \\ 
\addlinespace[3pt]
Traffic incident   & \cite{tran2023msgnn}   & MSGNN & ChebyGIN  & K-POD clustering  & Vehicle crashes, etc. & Australia  & Subarea level incident &   
\\ 
\addlinespace[3pt]
Pavement performance  & \cite{cai2023fine}  & CTGCN  & GCN   & LSTM  & Pavement images, etc.  & Shanghai  & Causal graphs  &  \\
\addlinespace[3pt]
Truck loan & \cite{chen2024spatial}  & SGTD  & Gated GNN  & LSTM  & Truck GPS, loan records  & China  & Spatio-temporal  &   \\
\addlinespace[3pt]
\bottomrule[1.5pt]
\end{tabular}}
\end{sidewaystable}

\subsubsection{Spatial modules: GNNs}
We now summarize GNN variants used in traffic prediction in Table~\ref{traffic_prediction}. Over half of the listed studies (16 out of 31) developed their models using GCNs as foundation architectures. Recall from Section~\ref{gnn_model} that GCNs employ matrix convolutional operations for information propagation, generally providing higher computational efficiency than the neighborhood aggregation in GraphSAGE~\citep{liu2020graphsage} and the edge-by-edge attention computation in GATs. By combining graph convolution operations in GCNs and recurrent units, MegaCRN captures traffic variability across sensors and periods, as well as irregular disruptions caused by accidents~\citep{jiang2023spatio}.

Alternatively, the NRI local, NRI unif, and NRI DTW models, designed for traffic speed and taxi demand prediction~\citep{tygesen2023unboxing}, were built on the Graph Network~\citep{battaglia2018relational}. Note that the Graph Network updates information directly utilizing nodes, edges, and global graph characteristics, thereby constituting a generalized extension of GCNs. Furthermore, ChebyGIN~\citep{knyazev2019understanding}, which extends GCNs by considering multi-hop neighborhoods within one iteration, acts as the base in MSGNN for traffic incident prediction~\citep{tran2023msgnn}. 



\subsubsection{Temporal modules}
GNNs can effectively capture spatial dependencies in traffic prediction tasks. However, additional components are still necessary to model temporal relationships. Table~\ref{traffic_prediction} indicates that temporal modeling techniques, including the Long short-term memory (LSTM)~\citep{hochreiter1997long}, GRUs, and Transformers with attention mechanisms~\citep{vaswani2017attention} were frequently integrated with GNNs. These temporal modules maintain strong capabilities to propagate information along the temporal axis. 

\subsubsection{Datasets}
Many GNN-based traffic prediction studies predominantly used traffic speed and flow rate data from highway traffic datasets from California, such as PEMS-BAY, METR-LA~\citep{li2017diffusion}, PeMSD4, and PeMSD8~\citep{guo2019attention}. These datasets consist of real-time traffic variable measurements collected from detector sensors, typically recorded at five-minute intervals. Recent research, however, has started to include various new data sources like taxis, trucks, accidents, electric vehicles, E-scooters, and flights (Table~\ref{traffic_prediction}). Besides, case studies have broadened to countries such as China~\citep{feng2023urban}, Australia~\citep{tran2023msgnn}, and South Korea~\citep{trirat2023mg}. These studies demonstrate GNNs' effectiveness in spatial modeling for diverse DMTN problems globally. Despite these advancements, the direct application in many developing countries remains limited. Their traffic data might have high variations due to mixed vehicle and pedestrian environments, which poses challenges for diverse traffic prediction tasks.

\subsubsection{Interpretability}

Existing interpretation techniques for GNNs in traffic prediction originate from general approaches summarized in Section~\ref{general}. In particular, \cite{huang2024explainable} built GNN models that account for delay propagation patterns in railway systems. This study, which falls within the feature-level category, identifies train headway as the critical factor for train delay propagation. Furthermore, \cite{tygesen2023unboxing} conducted interoperability analysis by examining learnable adjacency matrices that reflect internodal interdependencies. Similarly, the collaborative prediction unit methods proposed by \cite{li2022online} explained GNNs from an edge perspective. Accordingly, the above two studies are graph-level. Inspired by the GNNExplainer~\citep{ying2019gnnexplainer}, Traffexplainer unveils critical spatial and temporal elements in traffic prediction through iterative updates of masked spatial and temporal matrices. Consequently, Traffexplainer belongs to both the feature-level and graph-level interpretation methods~\citep{kong2024traffexplainer}. Note that the interpretation approaches for GNNs are closely associated with adversarial attacks on GNNs, which involve modifying a subset of nodes within transportation networks to harm traffic prediction performance~\citep{zhu2023adversarial}. Such analyses can contribute to developing robust GNN prediction models that maintain high performance across diverse contexts.

In summary, the applications of GNNs in traffic prediction exhibit three key trends: (1) from roadway traffic to multiple transportation modalities, (2) from standard GNN architectures to customized modules designed to capture stochastic, human-centric, and multi-resolution facets of traffic dynamics, and (3) from limited data sources to diverse, multinational datasets.

\subsection{Traffic operation}
\label{traffic_optimization}
\begin{figure}[h]
    \centering
    \includegraphics[scale=0.68]{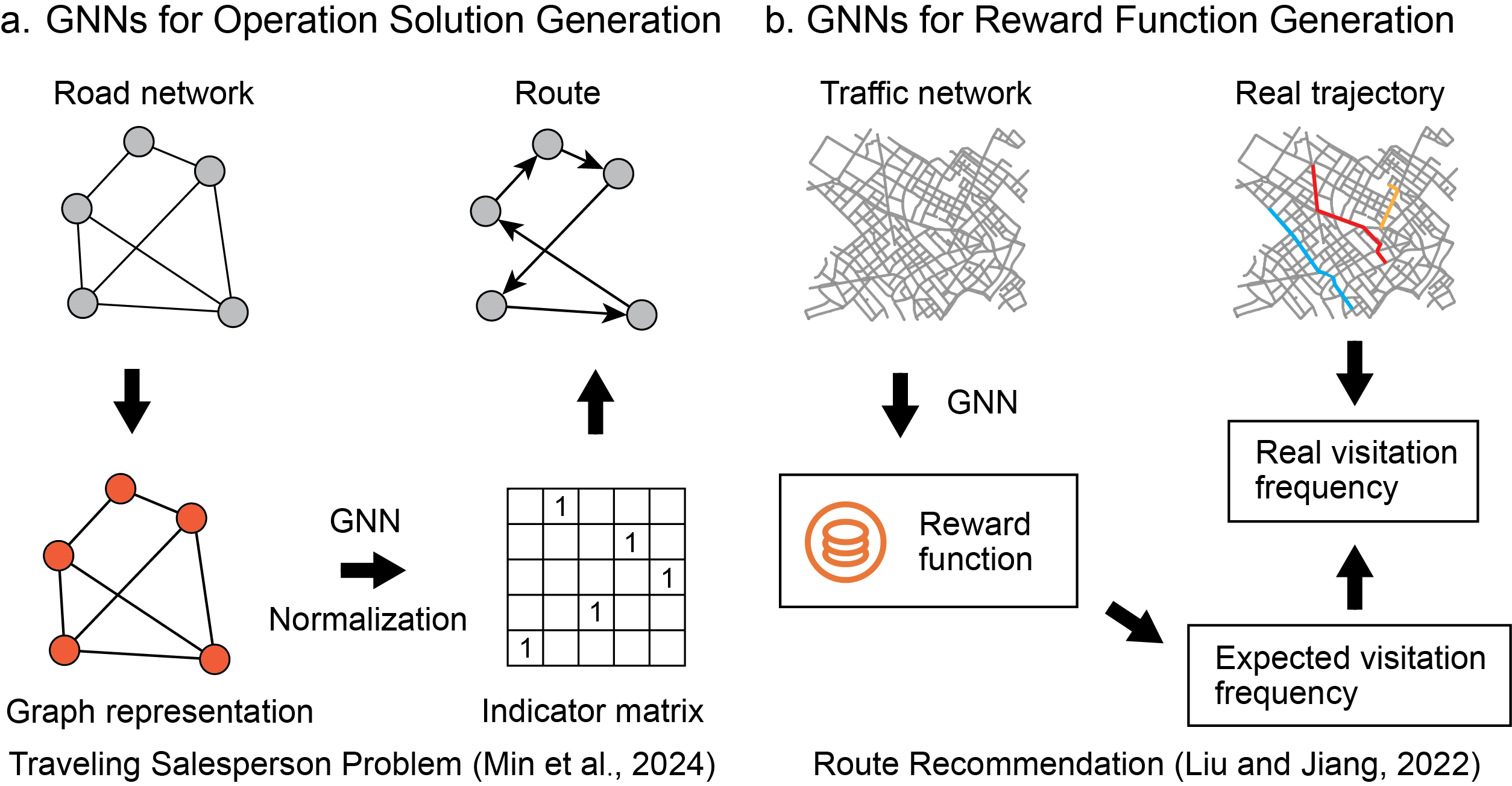}
    \caption{Applications of GNNs in traffic operation. (a) GNNs can generate route solutions by incorporating road network topology with well-designed loss functions~\citep{min2024unsupervised}. (b) GNNs act as reward function generators for route optimization~\citep{liu2022personalized}.}
    \label{fig_operation}
\end{figure}

GNNs are capable of learning informative representations for nodes and edges that encode graph structures. These learned representations can be utilized to derive solutions for graph-based operation problems. For instance, computational scientists have utilized GNNs to address combinatorial optimization problems such as general mixed-integer linear programs~\citep{lee2024rl}, the Maximum Cut~\citep{heydaribeni2024distributed}, the Maximum Independent Set~\citep{schuetz2022combinatorial}, the Traveling Salesperson Problem (TSP)~\citep{min2024unsupervised} (Fig.~\ref{fig_operation}a). In particular, they first employed GNNs to obtain continuous node representations based on the operation problem. After that, they either mapped continuous node variable values to integers~\citep{schuetz2022combinatorial} or used tree search methods to derive discrete solutions~\citep{min2024unsupervised}. Overall, GNNs can assist in addressing optimization problems by providing superior feasible solutions or evaluating the optimality of solutions~\citep{cappart2023combinatorial}. Alternatively, the spatial learning capabilities of GNNs enable their potential for addressing resource allocation problems, which determines the optimal distribution of limited resources within interconnected systems. For instance, \cite{wang2022learning} proposed Aggregation GNNs to utilize asynchronous and delayed signal information to optimize decentralized resource allocation in wireless networks.

In transportation networks, researchers have applied GNNs to address various traffic operation problems including vehicle repositioning~\citep{chang2022cooperative}, vehicle routing~\citep{liu2022personalized}, railcar itinerary optimization~\citep{zhang2025railcar}, and the control of connected autonomous traffic~\citep{zhou2024reasoning}. For example, an attention-enhanced GCN was developed to predict spatial travel demand for carsharing systems~\citep{chang2022cooperative}. The prediction outcomes acted as the input for downstream vehicle location optimization for profit maximization in carsharing companies. Besides, GNNs have been used in route recommendations~\citep{liu2022personalized} (Fig.~\ref{fig_operation}b) and the routing of unmanned aerial vehicles~\citep{fang2023routing}. In these applications, GNNs provide reward function values or new information during the interactive optimization process, leveraging their capability to capture the spatial-temporal correlations of traffic features within networks. For delivery service route planning, researchers used GNNs and Transformers to generate waypoints along the route and employed a divide-and-conquer method to determine detailed routes~\citep{zhang2023route}. 

Notably, GNNs have been leveraged to address the traffic assignment problem (TAP): given traffic demand in a road network, determine how traffic flows through the network~\citep{nguyen1974algorithm,daganzo1977stochastic}. In particular, ~\cite{rahman2023data} solved the TAP by learning a mapping from the demand matrix and the network structure to the link flow matrix using GNNs. They employed synthetic traffic data in the Sioux Falls network and a road network in Massachusetts from user equilibrium to train the GNN model, demonstrating that the GNN model was able to generate link flows with less than 2\% errors. Besides, ~\cite{liu2023end} created both real links and virtual links (which connect origin and destination nodes) in GNNs for solving the TAP. For model training, the authors customized a total loss function (i.e., $L$) as the weighted sum of three components: the link-level flow-capacity ratio error (i.e., $L_{f-c}$), the link flow error (i.e., $L_{f}$), and the node-based flow conservation error (i.e., $L_{c}$):
\begin{equation}
    L = \alpha L_{f-c} + \beta L_{f} + \gamma L_{c}.
\end{equation}

It is worth mentioning that the applications of GNNs in traffic operation problems are still in their early stages. The circumstances under which GNN solutions surpass existing methods in terms of running time and solution quality, for both large and small-scale networks, remain an open question~\citep{boettcher2023inability}. Current experiments of GNNs on the TAP are limited by networks with at most hundreds of nodes~\citep{rahman2023data,liu2023end}. It is unclear whether the successful online deployments of GNNs in large social networks, such as LinkedIn Graph with billion-level nodes~\citep{borisyuk2024lignn} (Section~\ref{gnn_variant}) can be effectively transferred to road traffic networks. Future research can explore this area further. 


\subsection{Accident prediction}
GNNs serve as novel approaches for traffic accident prediction, which plays a central role in traffic safety management~\citep{mannering2020big}. Two primary factors contribute to these applications. First, the robust learning capabilities of GNN neurons guarantee the extraction of complex relationships between traffic accident occurrences, road network profiles, and external factors from large-scale datasets, surpassing the limitations of traditional statistical methods and causal inference models in prediction accuracy. Second, the internode message-passing mechanism inherent in GNNs seamlessly fits the spatial interconnectedness of traffic accident events, which arise from traffic dynamics and naturalistic driving behaviors~\citep{sae2023spatial}. 

In particular, \cite{yu2021deep} developed a deep spatio-temporal GCN using diverse data sources (including traffic speed data, road network data, and weather data) for accident prediction, and obtained superior prediction performance over previous methods. Recently, transportation researchers applied GCNs and Latent Dirichlet Allocation~\citep{porteous2008fast} to examine traffic accident-related factors, such as daily vehicle kilometers traveled and residential activities using floating vehicle trajectory data in San Francisco~\citep{zhao2024exploring}. Crucially, traffic accidents display zero-inflated properties, meaning that many road segments report no accidents~\citep{dong2014multivariate}. To address this, researchers developed probabilistic GNN models to forecast accident risk distributions in London, which is particularly useful for road segments with varying risk levels~\citep{gao2024uncertainty}.





\subsection{Industry practice}
\label{industry_models}
Many technology companies have created specialized GNN models in online transportation services. A key application is the travel time estimation on road networks, which predicts arrival time based on departure time and routes (Table~\ref{industry}). This serves as a fundamental functionality form in digital maps like Google Maps. We now compare industry approaches with academic GNN research, looking at how they build graphs, design model architectures, and apply the technology.

Graph construction differs between academic research and industry applications. Influential academic studies such as DCRNN~\citep{li2017diffusion} and Graph WaveNet~\citep{wu2019graph} define nodes as stationary traffic sensors. In contrast, technology companies represent road segments as nodes~\citep{fang2020constgat,derrow2021eta}. Here, a road segment is a contiguous section of roadway, typically ranging from tens to hundreds of meters in length~\citep{dai2020hybrid}. This distinction leads to divergent prediction targets: academic studies focus on predicting speed and volume on sensor locations, while industry models aim to predict road segment-level travel time. Two major factors drive this distinction: 
\begin{itemize}
    \item Industry goals prioritize user-centric travel time estimation, which is more directly derived from road segment-based other than sensor-based predictions.
    \item Road segment-level travel time prediction requires traffic data on every road within road networks, which is generally accessible to technology companies, not academic researchers. 
\end{itemize}  

\begin{table}[htp]
\caption{\label{industry} GNNs in travel time estimation from industry. Target: the entities for which travel time predictions are being performed. APE: absolute percentage error. MAE: mean absolute error. MAPE: mean absolute percentage error. RMSE: root mean squared error.}
\resizebox{1.00\textwidth}{29.0mm}{
\begin{tabular}{@{}llll@{}}
 \toprule[1.5pt]
  \addlinespace[4pt]
Study  & \cite{derrow2021eta}  & \cite{dai2020hybrid}  & \cite{fang2020constgat} \\ 
 \addlinespace[3pt]
 \toprule[1.5pt]
  \addlinespace[3pt]
Model  & GN  & H-STGCN  & ConSTGAT \\ 
 \addlinespace[3pt]
Company   & Google   & Amap  & Baidu  \\
 \addlinespace[3pt]
Target & Supersegment & Road segment   & Road segment  \\
 \addlinespace[3pt]
Node  & Road segment  & Road segment   & Road segment  \\
 \addlinespace[3pt]
Edge   & Adjacent  & Any two   & Adjacent  \\
 \addlinespace[3pt]
Base module  & Graph Network~\citep{battaglia2018relational}  & GCN~\citep{defferrard2016convolutional}  & 3D-attention on graphs  \\
 \addlinespace[3pt]
Other modules  & MetaGradient~\citep{xu2018meta} & Temporal gated convolution~\citep{yu2017spatio} & Segment-based methods  \\
 \addlinespace[3pt]
Loss function  & Huber loss  & L1 loss  & Huber loss, APE  \\
 \addlinespace[3pt]
Area  & New York, Los Angeles, Tokyo, Singapore & Beijing & Taiyuan, Hefei, Huizhou \\
 \addlinespace[3pt]
Performance  & RMSE  & MAE, RMSE, MAPE  & MAE, RMSE, MAPE   \\
 \addlinespace[3pt]
Deployment & Yes & Not mentioned   & Yes  \\ 
 \addlinespace[1.5pt]
\bottomrule[1.5pt]
\end{tabular}}
\end{table}

Industrial GNN applications in transportation were generally built upon established GNN architectures (Table~\ref{industry}). For instance, Google's ETA prediction utilized the Graph Network~\citep{battaglia2018relational}, which incorporated convolutions on nodes, edges, and the entire graph. Alternatively, Alibaba resorted to Chebyshev polynomials of graph Laplacian for graph convolution operations~\citep{defferrard2016convolutional}. Notably, Baidu's approach, while not explicitly using adjacency matrices for representation updates, integrated network structure information via tensors and applied three-dimensional attention mechanisms from Transformer architectures. To meet low-latency requirements in service, they created look-up tables for swift prediction outcome retrieval.

GNN models for travel time estimation from industry have been applied to diverse cities globally (Table~\ref{industry}). Google conducted offline case studies using traffic data from New York, Los Angeles, Tokyo, and Singapore, and online evaluations in cities from North America, Europe, Asia, and the Pacific region~\citep{derrow2021eta}. Besides, Amap and Baidu implemented their methods using crowdsourcing data from their applications in various Chinese cities~\citep{dai2020hybrid,fang2020constgat}. While academic research employed MAE, RMSE, and MAPE to evaluate model prediction accuracy, industry practitioners have also incorporated negative ETA outcomes, which measure the events when the ETA errors exceed specified thresholds~\citep{derrow2021eta}. It prioritizes large prediction errors and ignores minor gaps, potentially contributing to improved user satisfaction ratings for their services.

Together, the implementations of GNNs in transportation networks by technology companies are sometimes grounded in academic research and address various deployment challenges. Many industrial travel time estimation products are fundamentally based on GCNs from academic studies. While academic experiments are constrained by limited open datasets and less practical sensor-level predictions, industry applications have expanded to road segment-level prediction across global cities. This process involves advanced training approaches such as  MetaGradient~\citep{xu2018meta} and computational methods such as parallel computing to enhance user experience. However, due to privacy concerns, industry studies have not disclosed their codes and data, creating obstacles for academic researchers to further refine their methodologies.

%% file: opportunities_gnn.tex
\section{Future Opportunities
 of GNNs in Transportation Problems}
\label{opportunies}
\begin{figure}[h]
    \centering
    \includegraphics[scale=0.68]{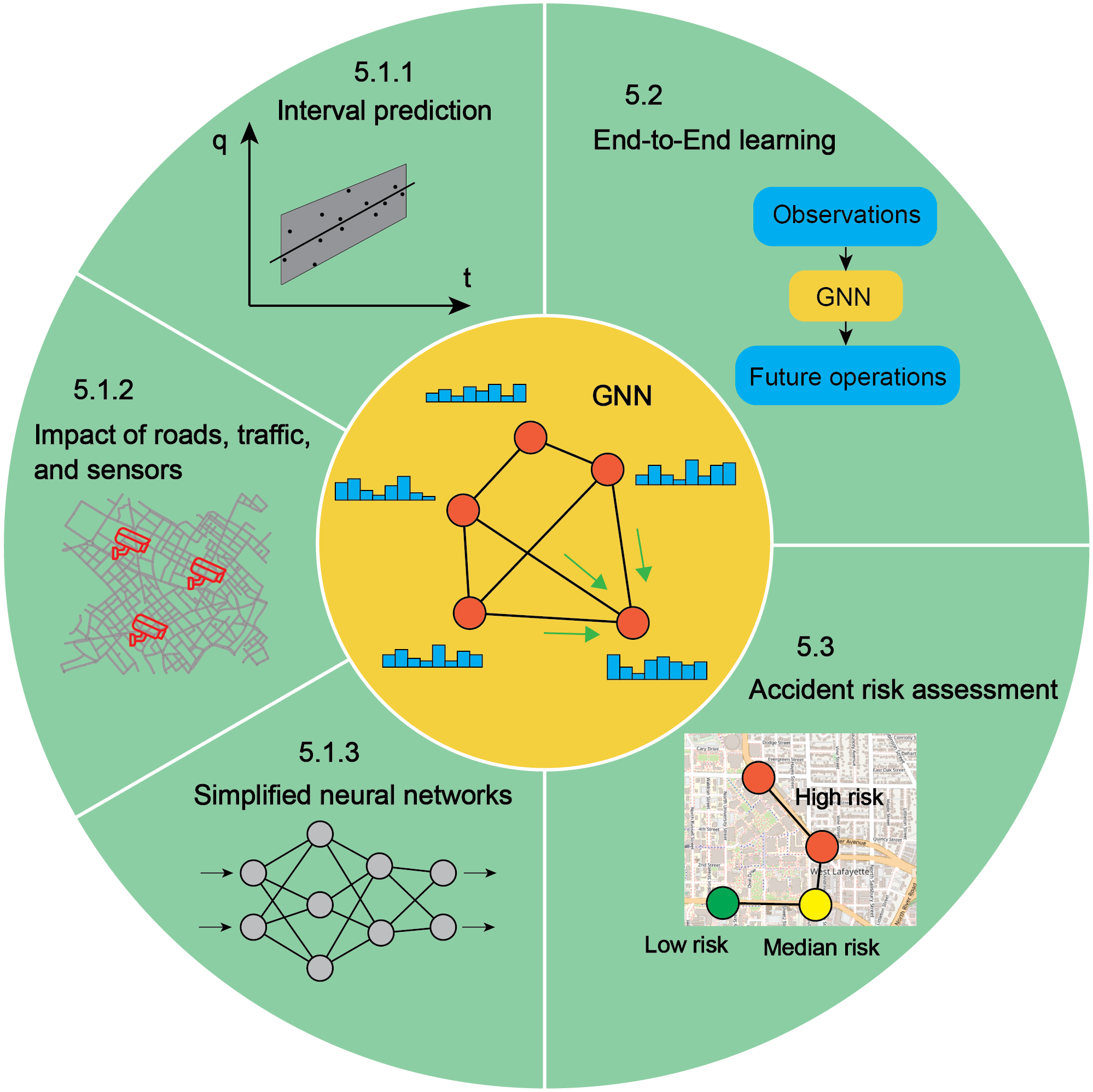}
    \caption{Prospective directions of GNNs in transportation. These include GNNs for traffic prediction (from Section 5.1.1 to Section 5.1.3), traffic operation (Section 5.2), and accident mining (Section 5.3). Background map\;\textcopyright\;OpenStreetMap.}
    \label{fig_future}
\end{figure}

This section provides an overview of future directions of GNNs in various transportation problems (Fig.~\ref{fig_future}). The deployment of GNNs in transportation can become more robust, interpretable, and efficient across a wider range of transportation problems. This trend will be propelled by key factors such as (1) increasingly accessible traffic datasets, (2) advancements in fundamental neural network architectures, and (3) extensive interactions between GNNs and related fields such as operation research.  

\subsection{Traffic prediction}
\subsubsection{Moving from single-point prediction to interval prediction}

Despite significant progress in traffic prediction methodologies, there has been limited research on traffic uncertainty. Traffic uncertainty refers to the variability in traffic conditions affected by multiple factors such as special events and abrupt weather changes. Under uncertain circumstances, actual traffic demand and travel time might exceed anticipated values, leading to inefficiencies in traffic management such as navigation for emergency vehicles like ambulances. To account for this uncertainty, a viable approach is to provide prediction intervals or distributions~\citep{chen2023probabilistic,xu2024link}, as opposed to single-point estimates. A recent study has developed a probabilistic GNN to forecast travel demand intervals in rail systems and the ridesharing market~\citep{wang2024uncertainty}. It utilized calibration errors, prediction interval width, and interval coverage probability as metrics for model evaluation. Analogously, a probabilistic GNN model was proposed to forecast sparse origin-destination flows in New York City~\citep{liang2024generating}. A promising direction is extending these models from travel demand forecasting to roadway traffic condition prediction. 

Moreover, the determinants affecting the length of predicted confidence intervals for traffic values, which measure prediction uncertainty, remain unexplored. Given the capability of GNNs to learn general patterns in training datasets, we posit that the statistical dispersion of traffic variables within training sets may be a significant contributing factor to interval lengths.

\subsubsection{Investigating the effects of road types, traffic patterns, sensor placements, special scenarios}
Multiple critical factors of GNN-based traffic prediction models require systematic investigations. These include the effects of road types, traffic patterns, sensor placements, as well as special regions and time periods on prediction accuracy. These factors are critical for the successful transition of laboratory-developed GNN models to practical applications across diverse real-world urban, suburban, and rural areas. 

First, the extensively used datasets METR-LA and PEMS-BAY~\citep{li2017diffusion} focus on highways in California, with relatively homogeneous traffic conditions. It is insightful to extend the scope of their studies to cover urban streets, such as downtown streets in over 30 European cities from the UTD19 dataset~\citep{loder2019understanding}. We hypothesize that GNN-based traffic prediction models applied to urban streets may exhibit higher relative error rates than highway applications, assuming consistent external conditions. This is because urban streets have more complex and heterogeneous traffic patterns including interactions between pedestrians, cyclists, and dense traffic control signals.

Second, the relationships between traffic patterns and prediction accuracy remain inadequately understood. There is little evidence to test the hypothesis that increased traffic volumes consistently result in larger prediction errors. The recently introduced LargeST dataset~\citep{liu2024largest}, which includes traffic data from thousands of sensors with diverse traffic profiles, presents a versatile opportunity to solve this problem. Additionally, many prediction approaches performed the Z-score normalization~\citep{wu2019graph}, which standardizes data values to a mean of zero and a standard deviation of one. This data preprocessing method might mitigate the influence of traffic volumes on prediction outcomes. 

Third, there is no consensus on the influence of sensor placements on GNN-based traffic prediction outcomes. Sensor locations directly influence adjacency matrices in graph convolution operations, affecting the learning process and prediction results. A study demonstrated that the traffic flow prediction errors under the coarse scale were lower than those under the finer scale in Chicago~\citep{wang2022multivariate}. Alternatively, the relative errors for GNN-based traffic imputation were lower with the compact sensor sampling than the coarse sensor sampling in London~\citep{xue2024network}. Despite these observations, standardized criteria for characterizing sensor locations and assessing prediction accuracy are necessary before conducting systematic investigations of their effects.

Fourth, it is unclear whether current GNN-based traffic prediction approaches perform well in addressing special locations and temporal periods, which are crucial for dynamic traffic management. In particular, traffic patterns in road networks around large public venues (e.g., sports arenas and convention centers) exhibit large uncertainty due to complex traffic dynamics. Similarly, both intra-urban and inter-urban traffic conditions during special circumstances (e.g., city marathons and heavy rains) are also likely to diverge from regular traffic patterns because of factors like road closures\footnote{https://www.cbsnews.com/newyork/news/nyc-marathon-closures-road-bridge-street-2024/}. To achieve accurate predictions in these exceptional scenarios, GNN models need to learn from sparse historical data while adapting to real-world traffic fluctuations.

\subsubsection{Simplifying GNNs}
Section~\ref{gnn_model} illustrates that vanilla GNN models (like GCNs and GATs) were initially developed for general data mining tasks in networks including social and biological networks. However, their graph convolution operations may be redundant for traffic prediction applications, which necessitate low latency. Hence, a promising research direction is to simplify existing GNN methods to enhance traffic prediction efficiency. To achieve this, it is essential to identify key information extracted by GNNs that contributes to their superior performance over multilayer perceptions (MLPs) in prediction tasks. Recent research has found three key factors: sensor locations, time of day, and day of the week~\citep{shao2022spatial}. Following this, they came up with a simple but efficient MLP model that beats various GNN models in traffic prediction tasks. Besides, graph-less neural networks (GLNNs) have been proposed as mixed models, combining GNNs with high accuracy and MLPs with fast inference capabilities~\citep{zhanggraph}. GLNNs distill knowledge from GNNs and use the knowledge to train MLPs in an offline manner. The trained MLPs are subsequently exploited in online inference. Both studies provide a foundation for further simplification of existing GNN with increased traffic prediction efficiency.

\subsection{Traffic operation}
As discussed in Section~\ref{traffic_optimization}, researchers have utilized GNNs to predict carsharing demand and optimize profits~\citep{chang2022cooperative}. This framework is called \textit{Predict-then-Optimize} in operation research~\citep{elmachtoub2022smart}. An alternative approach is the \textit{End-to-End} (E2E) method, which directly derives optimization solutions from input features~\citep{qi2023practical,liu2023end}. While GNNs have demonstrated efficacy in specific \textit{Predict-then-Optimize} frameworks~\citep{chang2022cooperative}, their effects on E2E frameworks are still unexplored. Further research in this direction may enhance existing operation problems with spatial components.

For classical NP-hard graph problems (like the TSP problem), there are several benchmarking datasets (like National TSP instances\footnote{https://www.math.uwaterloo.ca/tsp/world/countries.html}). These datasets can strongly facilitate the development and evaluation of GNN-based operation methods on graphs. However, there is a shortage of standardized datasets and scenarios for diverse traffic control and management problems. Researchers have developed GNN-based control methods for connected autonomous vehicles~\citep{chen2021graph}, using customized scenarios generated by the traffic simulator SUMO~\citep{krajzewicz2012recent}. The creation of standardized testbeds for traffic operation problems would significantly promote GNN-based methods in this field.

\subsection{Accident prediction}
This subsection delineates two future directions for GNN-based traffic accident prediction. First, while existing GNN-based traffic accident predictions have primarily concentrated on overall crash occurrence~\citep{wu2023multi} and risk scores~\citep{gao2024uncertainty}, significant research opportunities persist in the exploration of accident types. These include the diversity in collision categories (e.g., rear-end collisions and single-vehicle accidents) and locations (e.g., urban streets, highways, and intersections). The extent to which spatial accident relationships, revealed by GNN models, vary across these accident categories remains an open question. Second, given the intrinsic interdependence between traffic flow dynamics and accident occurrence, the integration of traffic flow features (e.g., flow rate and traffic density) as side information within GNN-based accident prediction frameworks presents a promising research direction. Note that it is symmetrical to accident-aware traffic prediction, where accident occurrence is used as supplementary information for predicting traffic~\citep{ye2023dynamic}. The direct benefit is to enhance current accident forecasting accuracy. Besides, this discovery has the potential to further unveil complicated interconnections and causal mechanisms within traffic systems and accident landscapes~\citep{wang2013spatio,retallack2019current}.


\subsection{Integration with traffic theories}

As a drawback, the inference process of GNNs in transportation networks may not adhere to established traffic theories, resulting in limited trust from transportation engineers in GNN inference outcomes. A potential solution is the physics-informed GNN (PIGNN)~\citep{dalton2023physics,niresi2024physics,mo2024pi}, which fuses physics laws with GNNs to achieve both model interpretation and accuracy. PIGNN represents a subset of physics-informed machine learning (PINN)~\citep{karniadakis2021physics}, which embeds physics theories into machine learning to efficiently extract patterns from noisy data. The advantage of PINN has inspired many PINN transportation studies, including traffic flow~\citep{yuan2021macroscopic,shi2021physics} and car-following modeling~\citep{mo2021physics}. However, despite the emergence of PIGNN studies in transportation~\citep{zhu2022kst, xue2024network}, the optimal balance between physics-based and GNN components in PIGNN across various transportation contexts remains unclear. 

\subsection{Industry practice}
A potential direction for technology companies is the refinement of existing GNN-based travel time prediction methodologies. This process necessitates the incorporation of exterior factors, including weather records, traffic propagation patterns, and special event occurrences. For example, Baidu Maps has integrated congestion propagation patterns with graph learning techniques to improve travel time estimation~\citep{huang2022dueta}. The developed DuETA model, utilizing congestion-sensitive graphs and route-aware Transformers, has achieved superior prediction performance in major Chinese cities like Beijing, Shanghai, and Tianjin than the baseline model ConSTGAT (Table~\ref{industry}). Besides, transportation researchers have revealed a correlation between evening sporting events and reduced morning traffic congestion the next morning~\citep{yao2021twitter}. Technology companies can leverage diverse data sources, including event and foot traffic information, to enhance their intelligent transportation services.

Industry participants can also play a role in enhancing traffic safety management by running roadway crash risk assessment systems. The risk evaluation outcomes generated from GNN models, when communicated to drivers, may potentially reduce crash incidents in high-risk scenarios. The viability of this idea has been supported by a recent study that developed spatiotemporal GNN models for crash risk evaluation using multi-sourced data~\citep{liu2023attention}. The model addresses the inherent imbalance in traffic crash data (where non-crash data significantly outnumbers crash data) by utilizing focal loss~\citep{lin2017focal} during model training. While GNN-based crash risk evaluation remains a new field, it presents a promising chance for technology companies to develop robust and applicable risk-alert systems to secure roadway users.

Open data and code access is a significant future direction. While academic researchers like DCRNN~\citep{li2017diffusion} often make their data and code in GNN studies publicly available, many technology companies (like the three studies in Table~\ref{industry}) have not published their data or code. This creates giant barriers for researchers to address current limitations in GNN deployment within transportation systems. Although data privacy concerns may be major constraints for these companies, they could disseminate aggregated and non-sensitive data and code. Such effort would both satisfy the data-sharing needs of GNN communities and follow data confidentiality regulations.


%







%% file: data_code.tex
\section{Collection of Data and Code}
\label{data}

This section provides a summary of publicly available datasets, codes, and alternative learning resources tailored for GNNs and transportation networks. Given the existing reviews such as~\cite{shaygan2022traffic}, we
primarily focus on recent advances since late 2022.

\subsection{Datasets}
For traffic prediction, several high-quality transportation network datasets can be potentially utilized in future GNN studies (Table~\ref{dataset_2022}). Notably, the U.S. 20 dataset~\citep{xu2024unified} and LargeST~\citep{liu2024largest} include significantly more extensive areas compared to the PEMS-BAY and METR-LA~\citep{li2017diffusion}, which were confined to the Bay Area and Los Angeles. Here, the U.S. 20 dataset includes traffic flow, traffic density, and average speed data, while the LargeST contains traffic flow records. The U.S. 20 and LargeST datasets can facilitate the analysis of GNN-based traffic prediction across diverse road types and urban-rural settings within the United States. Beyond vehicle traffic, significant chances exist for traffic prediction in other transportation modes. Platforms like Flightradar24\footnote{www.flightradar24.com/} and MarineTraffic\footnote{www.marinetraffic.com/} offer real-time global movement data for commercial flights and marine vessels. These raw datasets enable us to develop customized GNN models to predict the dynamics of global traffic systems at various resolutions across routine operations and special events.  

For traffic operation, the growing penetration rate of sensing infrastructures has yielded substantial vehicle trajectory data from cities in the U.S.~\citep{xu2022highway}, China~\citep{yu2023city,wang2023city}, and Europe~\citep{neun2022traffic4cast}. Such datasets present opportunities for developing GNN-based vehicle routing algorithms.

\begin{table}[htp]
\caption{\label{dataset_2022} Open traffic datasets from late 2022.}
\resizebox{1.00\textwidth}{16.0mm}{
\begin{tabular}{@{}lllllr@{}}
 \toprule[1.5pt]
 \addlinespace[3pt]
Data & Type & Area & Temporal range & Spatial range  & Release date\\  
\addlinespace[3pt]
\toprule[1.5pt]
\addlinespace[3pt]
U.S. 20 dataset~\citep{xu2024unified} & Flow, density, speed & U.S. & 2019 & 20 cities & Mar. 2024  \\
\addlinespace[3pt]
LargeST~\citep{liu2024largest} & Flow  & California & 2017-2021 & 8600 sensors & Dec. 2023  \\
\addlinespace[3pt]
Camera data~\citep{yu2023city} & Vehicle trajectories & Shenzhen, Jinan & 2020-2022  & 1460 and 1838 sensors & Oct. 2023   \\
\addlinespace[3pt]
Vehicle identification~\citep{wang2023city}& Vehicle trajectories, flow & Xuancheng & 2020 & 80000 vehicles & Jan. 2023 \\
\addlinespace[3pt]
Traffic4cast~\citep{neun2022traffic4cast} & Vehicle counts & London, Madrid, Melbourne & 2019-2021 & 3751, 3840, 2589 detectors & Dec. 2022  \\
\addlinespace[3pt]
Fire evacuation~\citep{xu2022highway} & Vehicle trajectories  & California & 2019 & 21160 records & Oct. 2022 \\ 
\addlinespace[1pt]
\bottomrule[1.5pt]
\end{tabular}}
\end{table}


\subsection{Codes}
In contrast to the early period of GNN research, there is currently an abundance of code resources. For example, the PyTorch Geometric (PyG)\footnote{https://pytorch-geometric.readthedocs.io/en/latest/} serves as a user-friendly Python library on PyTorch for GNNs. In this context, PyG provides essential functionalities for GNN research, including graph construction, the design of convolutional mechanisms, and distributed training. Similarly, the Deep Graph Library (DGL)\footnote{https://docs.dgl.ai/index.html} enables users to run GNN models on various machine learning frameworks like PyTorch and TensorFlow. DGL incorporates a customized data structure, DGLGraph, to store and process graph information, which can be particularly beneficial for large-scale transportation network modeling. Additionally, certain open courses, such as CS224W\footnote{https://web.stanford.edu/class/cs224w/}, offer fundamental GNN code resources for learners. Furthermore, several GNN researchers have made their model codes publicly available, as outlined in Table~\ref{traffic_prediction}. 

\subsection{Others}
The GNN research community has been actively compiling recent publications in the field. Their collections can inspire further applications of GNNs in transportation network analysis. For instance, the repository \textit{Spatio-Temporal Prediction Papers}\footnote{https://github.com/uctb/ST-Paper} includes collections of GNN papers in diverse domains such as traffic forecasting~\citep{kong2024spatio}, human mobility analysis~\citep{jiang2023learning}, pandemic predcition~\citep{xue2022multiwave}, and road safety analysis~\citep{nippani2024graph}. Besides, a separate repository\footnote{https://github.com/jwwthu/GNN4Traffic} summarizes GNN studies on traffic forecasting spanning 2018 to 2024. Except for the above application studies, theoretical GNN papers have been categorized in another repository\footnote{https://github.com/jiaqingxie/Theories-of-Graph-Neural-Networks}. It includes GNN applications to NP-hard problems~\citep{toenshoff2021graph}, the expressive power of GNNs~\citep{balcilar2021analyzing}, and the combination of GNNs and differential equations~\citep{poli2019graph}, which has been successfully applied to traffic prediction~\citep{choi2022graph}. Transportation researchers may find valuable insights from tutorials including current advancements and challenges in large-scale GNNs presented at KDD 2023\footnote{https://sites.google.com/ncsu.edu/gnnkdd2023tutorial?pli=1}, and the properties and practice of GNNs delivered at AAAI 2025\footnote{https://gnn.seas.upenn.edu/aaai-2025/}. These resources can provide valuable insights into learning methods applicable to transportation networks.

%% file: conclusion.tex
\section{Conclusion}
\label{conclusion}

Deriving meaningful patterns and insights from diverse transportation network data supports a wide range of intelligent transportation tasks. The spatial correlations within transportation network data can be effectively captured by GNNs, enabling numerous GNN-based DMTN studies. In this paper, we provide a broad and timely overview of these applications, integrating perspectives from academic research and industry practice. Specifically, we begin by discussing fundamental GNN models (e.g., GCNs, GraphSAGE) and their subsequent variants (e.g., LightGCN). In the context of transportation networks, we review traffic prediction studies on traffic speed, flow, demand, and traffic operation topics such as vehicle routing. Expanding upon these studies, we identify emerging directions in traffic prediction, including probabilistic traffic prediction, factor impact analysis, and streamlined model architecture. Besides, the potential applications of GNNs in end-to-end learning for operation problems, accident risk evaluation, and the integration of machine learning and physics-based models are extensively examined. Furthermore, we delve into the involvement of technology companies in GNN-based travel time estimation, with particular attention to differences in network construction compared to academic research. To foster further academic engagement, we enumerate up-to-date learning resources, including datasets, code repositories, and various learning materials. Our review could promote the application of GNNs in addressing transportation network challenges in the context of emerging technologies.

%% file: ack.tex
\section{Author Contributions}

\textbf{Jiawei Xue}: Conceptualization, Methodology, Visualization, Writing – original draft.
\textbf{Ruichen Tan}: Conceptualization, Methodology, Writing – original draft.
\textbf{Jianzhu Ma}: Conceptualization, Methodology, Writing – review \& editing.
\textbf{Satish V. Ukkusuri}: Conceptualization, Methodology,  Writing – review \& editing.

\section{Declarations}
\textbf{Conflict of interest}. All authors declare no competing interests.